\RequirePackage{fix-cm}
\documentclass[twocolumn]{svjour3}          
\smartqed  
\usepackage{CJKutf8}

\usepackage{lmodern}
\usepackage{mdwlist}
\usepackage{url}
\usepackage[hidelinks,bookmarks=false]{hyperref}
\usepackage[official]{eurosym}
\usepackage{epsfig, times}
\usepackage{longtable}
\usepackage{fancyhdr}
\usepackage{algorithmic}
\usepackage{amsmath}

\usepackage{float}
\usepackage{verbatim}
\usepackage{afterpage}
\usepackage{tablefootnote}
\usepackage[toc,acronym,xindy]{glossaries}
\usepackage[colorinlistoftodos,prependcaption,textsize=tiny]{todonotes}

\usepackage{color}
\usepackage{amsfonts}
\usepackage{tabularx}
\usepackage[T1]{fontenc}
\usepackage[utf8]{inputenc}
\usepackage{graphicx}
\usepackage{adjustbox}
\usepackage{subfigure}
\usepackage{arabtex}
\usepackage{utf8}
\usepackage{gensymb}

\usepackage{epsfig}
\usepackage{wrapfig}
\usepackage[ruled]{algorithm2e}

\usepackage{multirow}
\usepackage{amssymb,amsmath,amsthm,enumitem}

\begin{document}
\setlength{\abovedisplayskip}{3pt}
\setlength{\belowdisplayskip}{3pt}
\title{Active Learning for Event Detection in Support of Disaster Analysis Applications}


\author{\mbox{Naina Said}         \and
        \mbox{Kashif Ahmad}         \and
        \mbox{Nicola Conci}         \and  
	    \mbox{Ala Al-Fuqaha}         \and
       }


\institute{University of Trento, Italy\\Hamad Bin Khalifa University \at
              \email{kahmad@hbku.edu.qa}           
}

\date{Received: date / Accepted: date}

\maketitle

\begin{abstract}

Disaster analysis in social media content is one of the interesting research domains having abundance of data. However, there is a lack of labeled data that can be used to train machine learning models for disaster analysis applications. Active learning is one of the possible solutions to such problem. To this aim, in this paper we propose and assess the efficacy of an active learning based framework for disaster analysis using images shared on social media outlets. Specifically, we analyze the performance of different active learning techniques employing several sampling and disagreement strategies. Moreover, we collect a large-scale dataset covering images from eight common types of natural disasters. The experimental results show that the use of active learning techniques for disaster analysis using images results in a performance comparable to that obtained using human annotated images, and could be used in frameworks for disaster analysis in images without tedious job of manual annotation.


\end{abstract}

\section{Introduction}
\section{Introduction}
\label{sec:introduction}
Natural disasters, such as floods and earthquakes, may cause significant loss in terms of human lives and property. In such situations, an instant access to relevant information may help with timely recovery efforts.  In recent years, social media outlets have been widely utilized to gather disaster related information \cite{ahmad2018social}. However, the use of social media content also comes with lots of challenges. One such challenge is filtering out irrelevant information. To this aim, several frameworks have been proposed in the recent literature that rely on different classification and feature extraction techniques. One of the requirements of classification applications is the availability of sufficient training samples. However, annotation of training samples is a tedious and time consuming job, which requires lots of efforts. 

One of the possible solutions to reduce human labor in data annotation is the use of active learning techniques. Active learning has been widely utilized in a wide range of application domains having large quantities of unlabeled data and less quantities of labeled data. Such domains include Natural Language Processing (NLP), multimedia analysis and remote sensing \cite{liu2019generative,sener2018active,tuia2011survey,ahmad2018event,zhang2019active}. Active learning techniques have been recently used with Convolutional Neural Networks (CNNs) and Long-short Term Memory (LSTM) based frameworks to improve their overall performance \cite{sener2018active,karlos2019investigating}. Disaster analysis is relatively a new application that still lacks large collections of labeled data \cite{said2019natural}. We believe it could benefit from active learning. 
   
In this paper, we study and analyze the efficacy of utilizing active learning techniques in disaster analysis in social media images by employing and evaluating the performance of different active learning techniques in terms of classification accuracy. We mainly focus on the most commonly used scenario of active learning, namely, pool-based sampling that fits well in our disaster analysis task. In pool-based sampling, samples are drawn from a pool of unlabeled images into the initial small labeled training set. Under the above mentioned settings, we rely on two most commonly used query techniques; namely, (i) uncertainty sampling and (ii) query by committee.  We further evaluate the performance of these techniques with different sampling and disagreement strategies. For uncertainty sampling, we employ three different sampling strategies; namely, least confidence (LC), margin sampling (MS) and entropy sampling (ES). On the other hand, for query by committee based active learning approach, we explore and evaluate the capabilities of this approach with three different disagreement strategies; namely, vote entropy (VE), consensus entropy (CE) and max disagreement (MD). Moreover, we analyze and evaluate the performance of these methods using different number of queries by including a single image in the training set from the unlabeled pool of images to analyze how quickly each of the methods attains maximum accuracy.  

To the best of our knowledge no prior works explored such detailed analysis of active learning techniques in the relative new domain of disaster analysis applications. Moreover, considering the lack of large-scale (in terms of images as well as the number of disaster types/classes covered) benchmark datasets in the domain, we also provide a benchmark dataset containing a large number of images from most common types of natural disasters, as detailed in Section \ref{sec:dataset}. 

The main contributions of this work are:
\begin{itemize}

    \item[(i)] Stemming from the fact that machine learning techniques are driven by training data and annotating large volumes of data is a tedious and time consuming job, we carry out an analysis and evaluation study of active learning techniques with diversified set of sampling/disagreement strategies in support of disaster analysis applications. 
         \item[(ii)]  Through the introduction of the active learning techniques, we demonstrate that comparable accuracy can be achieved with active learning without involving human annotators in the tedious job of annotating large training sets, and active learning could be used in disaster analysis frameworks to obtain better results in scenarios where less annotated data is available. 
         \item[(iii)] We also analyze and evaluate the performance of the methods using different numbers of queries/iterations, which helps to provide a baseline for future work in the domain. 
          \item[(iv)] We also provide a benchmark dataset for disaster analysis applications covering images from eight different types of natural disasters.
\end{itemize}

The rest of the paper is organized as follows: Section \ref{sec:related} discusses the related work. Section \ref{sec:background} provides the background and reviews concepts of the active learning techniques. In Section \ref{sec:techniques} and \ref{sec:dataset} provide details of the proposed methodology and dataset, respectively. The details of the experimental setup, experiments and  results are provided in Section \ref{sec:results}. Finally, Section \ref{sec:conclusion} concludes this study. 

\section{Related Work}
\label{sec:related}

In recent years, disaster analysis of images shared on social media outlets received great attention from the research community. Several interesting solutions relying on diversified sets of strategies have been proposed to effectively utilize the available information. A majority of the efforts in this regard rely on multi-modal information including visual features and meta-data comprised of textual, temporal and geo-location information \cite{said2019natural}. For instance, Benjamin et al. \cite{bischke2017detection} utilized the additional information available in the form of meta-data along with visual features extracted through an existing deep model; namely, AlexNet, pre-trained on ImageNet \cite{deng2009imagenet}. Both types of information are then evaluated individually and in combination with flood-related images obtained from social media. 
Similarly, the work in \cite{ahmad2018social} also demonstrates better results for visual features over textual and other information from meta-data in disaster analysis. 

The majority of the visual features based frameworks for disaster analysis rely on existing pre-trained models either as feature descriptors or the models are fine-tuned on disaster related images. To this aim, the existing models pre-trained on both ImageNet \cite{deng2009imagenet} and Places \cite{zhou2014learning} datasets have been employed. For instance, in \cite{alam2018processing}, an existing model; namely, VGGNet-16 \cite{simonyan2014very} pre-trained on ImageNet is fine-tuned on disaster related images for categorization of the images into different categories, such as informative and non-informative, damage severity and humanitarian categories. Ahmad et al. \cite{ahmad2018comparative} utilized existing models pre-trained on both ImageNet and Places dataset as feature descriptors both individually and in different combinations. The authors also evaluate the performance of several handcrafted visual features extracted. 

More recently, disaster analysis of images shared on social media has also been introduced as a sub-task in a benchmark competition; namely,  MediaEval\footnote{http://www.multimediaeval.org} for two consecutive years. In MediaEval-2017 \cite{bischke2017detection}, the task focused on the classification of social media imagery into flood-related and non-flooded images. On the other hand, the task in MediaEval-2018 \cite{bischke2018multimediasatellite} focused on the identification of passable and non-passable roads in social media images. Majority of the solutions proposed for the classification of images into flooded and non-flooded  categories in MediaEval-2017 relied on deep models (e.g., \cite{ahmad2018social,bischke2017detection,nogueira2017data,avgerinakis2017visual}). For instance, in \cite{ahmad2018social} an ensemble framework relying on several deep models used as feature descriptors has been proposed. Similar trend has been observed in MediaEval-2018 for the identification and classification of passable roads through information available on social media, where majority of the methods relied on ensembles of deep models (e.g., \cite{ahmad2019automatic,feng2018extraction,Zhao2018multimediasatellite,Anastasia2018multimediasatellite,Armin2018multimediasatellite,bischke2018multimediasatellite}). For instance, in \cite{ahmad2019automatic} multiple deep models were jointly utilized in an early, late and double fusion manner. 

In the literature, disaster analysis in images has been mostly treated as a supervised learning task where classification models are trained on training samples annotated with human annotators. Two benchmark datasets, namely DIRSM \cite{bischke2017multimedia} and FCSM \cite{bischke2018multimediasatellite}, have been mostly reported in the literature \cite{said2019natural}. The datasets provide a limited set of images, which are not sufficient to train deep models. Moreover, both datasets cover flood related images, only. We believe active learning techniques could be useful to cover the limitation of lack of sufficient annotated training data in the domain. 

\section{Active learning: definitions and concepts} 
\label{sec:background}
Active learning is a semi-supervised learning technique which selects the training data it wants to learn from \cite{zhang2019active}. Selecting good training samples from the data enables active learning techniques to perform significantly better with fewer training samples compared to passive learning methods \cite{kading2018active}. In passive learning methods, a large chuck of the data is randomly collected from an underlying distribution for training purposes. The main advantage of active learning over passive learning is the ability to make a decision on the basis of the responses from the previous queries for choosing instances from the unlabelled pool of images. In this work, we mainly rely on pool-based sampling methods where samples are drawn from a large pool of unlabelled samples; namely, $u=\{x_i\}_{i=1}^{n}$. An initial training set also known as the seed denoted as $\upsilon=\{x'_i\}_{i=1}^{n'}$ is used the train the initial model, $\theta$, and is populated by picking and annotating the instances with $y_i=\{y_1,y_2, ... m\}$ from the unlabelled pool of samples, iteratively. 

In the next subsections, we provide a detailed description of the two query techniques (i.e., active learning schemes) used in this work along with the different sampling and disagreement methods used by those methods. 

\subsection{Uncertainty Sampling}

Uncertainty Sampling is one of the most common and widely used active learning techniques. With uncertainty sampling, the active learner queries the most uncertain instances (i.e., the samples for which the learner is least certain how to label). The technique is called uncertainty sampling because of its use of posterior probabilities in making decision, and is often straight forward for probabilistic learning models. For example, in case of binary classification, uncertainty sampling techniques simply ask for the instance that has a posterior probability of being positive around 0.5. For the selection of the samples, we employed several variants of this technique based on the informativeness measure of the unlabelled instances with three different sampling strategies; namely, (i) least confidence, (ii) margin sampling and (iii) entropy sampling. Next, we provide detailed description of those sampling strategies.

\subsubsection{\textit{Least Confidence Query Strategy}}

\begin{figure}[h]
\centering
\includegraphics[width=0.85\linewidth]{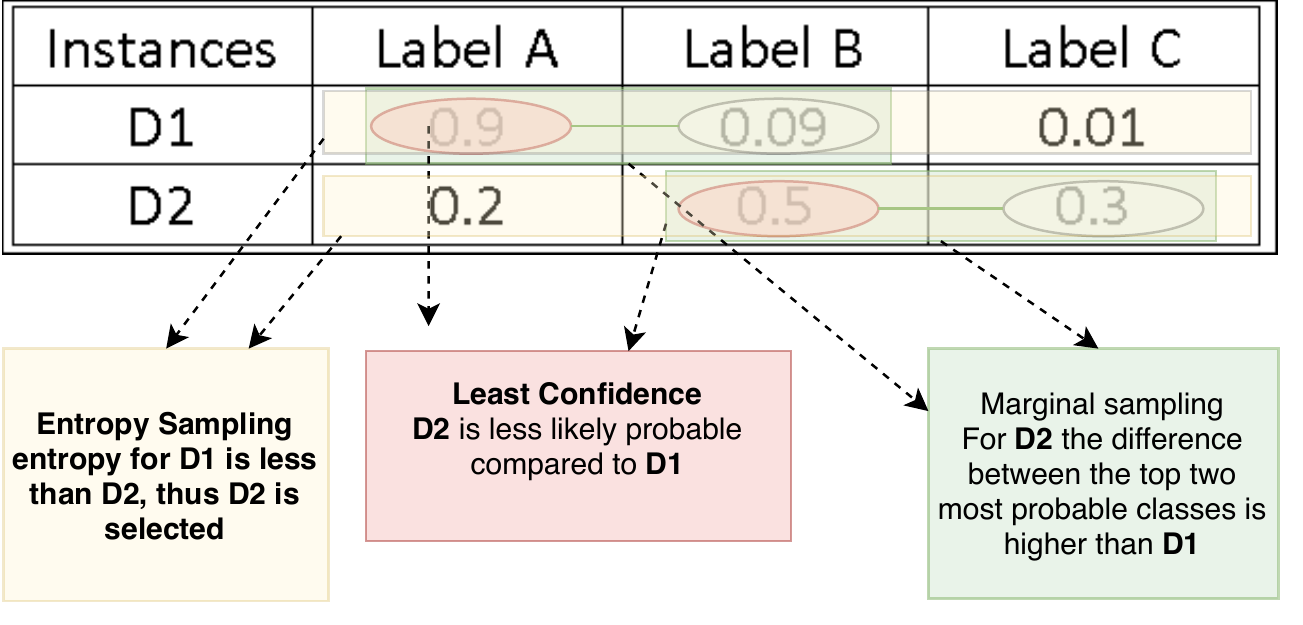}
\caption{An illustration of the working mechanism of the different sampling strategies used for uncertainty sampling. The sampling strategies; namely, LC, MS and ES, are represented in red, green and yellow colors, respectively. LC and MS consider the top 1 and 2 most probable labels while ES decides on the basis of the complete probability distribution considering all classes.}
	\label{fig:theme_2}
\end{figure}

This sampling strategy aims to choose the instance from the pool for which the learner has the least confidence about its most likely label as shown by equation \ref{equ:LC}, where $x$, $y^{'}$ and $\theta$ represent the sample, the most probably label and the underlying model, respectively. The strategy is more suitable for multi-class classification.  For example, if we have two unlabeled instances; namely, D1 and D2, having probabilities (p1, p2 and p3) with values (0.9,0.09,0.01) and (0.2,0.5,0.3) for class labels A, B and C, respectively, the Least Confidence (LC) query strategy selects D2 to be labeled as the learner is less sure about its most likely label. This example is illustrated in Figure \ref{fig:theme_2}. One way to interpret this query strategy is that the model selects an instance believed to be mislabeled. 

\begin{equation}
    \small
\centering
LC(X)= argmax_{x} 1 - p_{\theta} (y^{'} | x)
    \label{equ:LC}
\end{equation}

\subsubsection{\textit{Margin Sampling}}

One shortcoming of the LC query strategy is the decision on the basis of the most probable label only. The LC query strategy does not consider the rest of the labels which might be useful in the selection process. In order to cope with this limitation, Margin Sampling (MS) incorporates the posterior probability of the second most likely label by selecting an instance having the least difference between the top two most probable labels. Let's suppose $ y_1^{'}$ and $ y_2^{'}$ are the top two most probable labels for a sample $ x$ under a model $ \theta $. Then the margin between the two samples can be represented by equation \ref{equ:MS}. 

Considering the previous example presented in Figure \ref{fig:theme_2}, margin sampling selects D2 as the difference between its two most probable labels (i.e., $0.5 - 0.3 = 0.2$) is less than the difference between the two most probable labels of D1 (i.e., $0.9 - 0.09 = 0.81$). The low difference between the labels of D2 indicates that the instance is ambiguous and thus getting the true label of the instance would help in the classification process. 

\begin{equation}
  \centering
MS(X)= p_{\theta} (y1^{'} | x) - p_{\theta} (y2^{'} | x)
    \label{equ:MS}
\end{equation}

\subsubsection{\textit{Entropy Sampling}}

MS considers the top two most probable labels in the decision making process; however, for a dataset with higher number of class labels, the top two most probable labels are not sufficient to represent the probability distribution. To this aim, the Entropy Sampling (ES) strategy efficiently utilizes the probability distribution by calculating the entropy of each instance using equation \ref{equ:etropy}, where $P(y|x)$ represents the posterior probability while $H$ is the uncertainty measure and $Y$ is the output  class. Subsequently, an instance with the highest value is queried. In case of our example shown in Figure \ref{fig:theme_2}, D1 yields a value of 0.155 while D2 has a value of 0.447. Therefore entropy sampling selects the instance D2 for labelling. In case of binary classification, entropy sampling performs as margin and least confident sampling. However, it is most useful for probabilistic multi-class classification problems.

\begin{equation}
\small
\centering
ES(x)= -\sum_{y\epsilon{Y}}P_{\theta}(y|x)\log_{2} P_{\theta}(y|x)
    \label{equ:etropy}
\end{equation}

\subsection{Query By Committee}

The other active learning technique employed in this work is based on the query by Committee strategy. In this method, a query of different competing hypotheses (i.e., trained classifiers represented as $C={( \theta_{1}, \theta_{2}, \theta_{3} . . . \theta_{n}})$  of the current labelled data set namely $\lambda$ is maintained. The queries are then selected by measuring the disagreement between these hypotheses. The aim of the query by committee strategy is to reduce the version space, which is the set of hypotheses consistent with the current labelled set. For example, if machine learning is used to search for the best model within the version space then the aim of the query by committee method is to constrain the size of this space as much as possible leading to a more precise search with as few labelled instances as possible \cite{settles2009active}. In case of several hypotheses, the instance to be labeled next is chosen by measuring the disagreement among the hypotheses. Different strategies can be utilized to measure the disagreement, in this study we use three different strategies as detailed below. 

\subsubsection{Vote Entropy}

Vote entropy can be considered as query by Committee generalization of the entropy based uncertainty sampling, and is calculated by equation \ref{equ:VE}, where $y_{i} $ is the vector of all possible labels, $C$ represents the committee of the classifiers while $V(y_{i})$ is the total number of votes for label $y_{'}$. Suppose there are three classifiers (i.e., committee size is 3), three classes [0,1,2] and five unlabeled instances. Then, in order to calculate the vote entropy, every classifier is first asked for its prediction for all the unlabelled instances. Suppose the predictions returned for a single instance by all the three classifiers is [0, 1, 0] (i.e., classifier 1 predicts that the instance lies in class-0, classifier 2 predicts it as a sample from class-1 and classifier 3 also predicts it as class-0). Each instance has a corresponding probability distribution (i.e., the distribution of class labels when picking the classifier at random). In the stated example, there are two votes for 0, one vote for 1 and 0 votes for 2. Therefore, the probability distribution for this instance is [0. 6666, 0.3333, 0]. Among all the five instances, vote entropy selects the instance which has the largest entropy of this vote distribution. 

\begin{equation}
\small
\centering
VE(x)= arg_{x}max - \sum_{i} \dfrac{V(y_{i})}{C} \log \dfrac{V(y_{i})}{C}
    \label{equ:VE}
\end{equation}

\subsubsection{Consensus Entropy}

In consensus entropy, instead of calculating the probability distribution of the votes, the average of the class probabilities provided by each classifier in the committee is calculated. This average class probability is called the consensus probability. Once the consensus probability is calculated using equation \ref{equ:CE} (where $C$ represents the committee of the classifiers), its entropy is computed and the instance with the largest entropy is selected to be labelled by the labeler. 

\begin{equation}
\small
\centering
CE(x)= \dfrac{1}{C}\sum_{c = 1}^{C}P_{\theta}(y_{i})
    \label{equ:CE}
\end{equation}

\subsubsection{Max disagreement}

The Max disagreement sampling technique calculates the disagreement of each learner by using the consensus probability and then selects the instance with the largest disagreement. In this way, it deals with the issue of the other two strategies which take the actual disagreement into account in a weak sense.

\section{Methodology}
\label{sec:techniques}
\begin{figure*}[!htbp]
\centering
\includegraphics[width=0.78\linewidth]{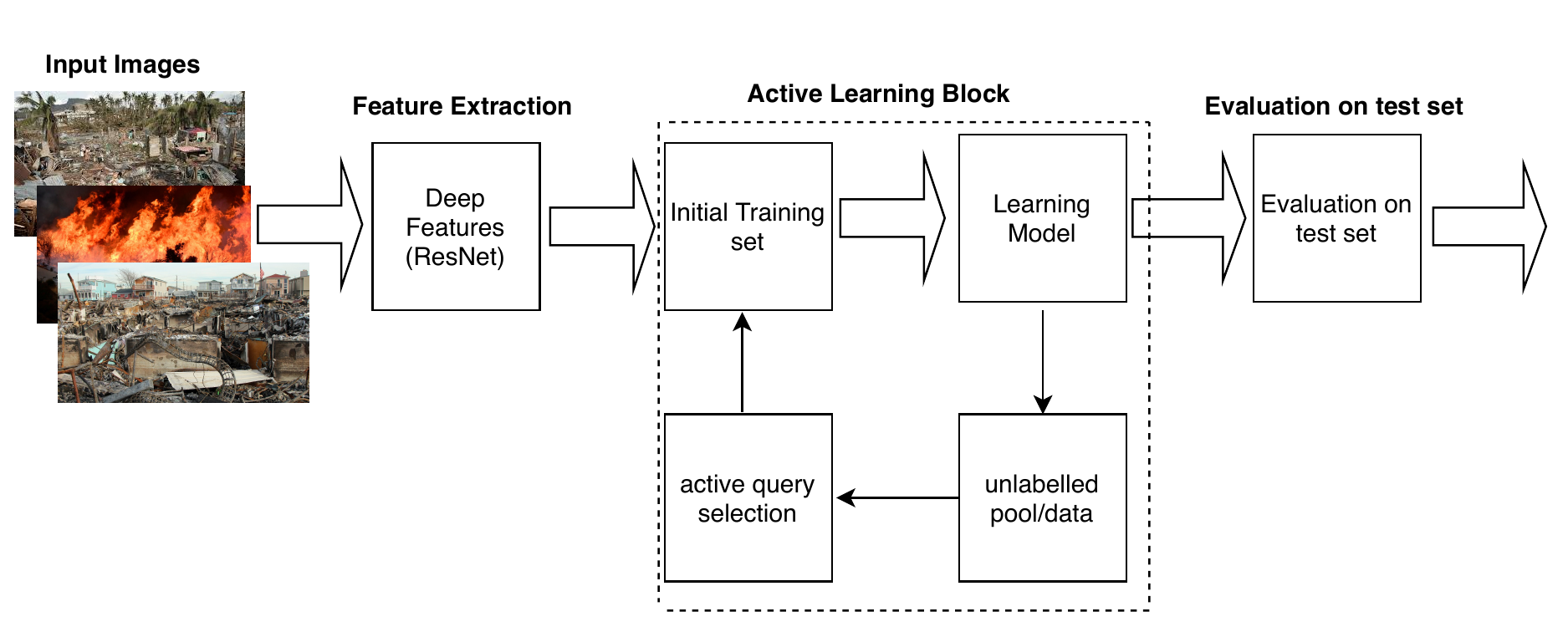}
\caption{Block diagram of the proposed methodology. 
}
	\label{fig:methodology}
\end{figure*}

Figure \ref{fig:methodology} provides the block diagram of the proposed methodology. The framework is composed of three main components; namely, (i) feature extraction, (ii) collection/annotation of the training samples through active learning and (iii) classification/evaluation. For feature extraction, we rely on an existing pre-trained model. For collection/annotation of the training samples, several active learning techniques are utilized. The classification phase is based on Support Vector Machines (SVMs). The feature extraction and classification phases are rather standard, and the main strength of the proposed framework stems from the active learning part where we collect/annotate relevant training samples from an unlabeled pool of images retrieved from social media outlets. In the next subsections, we provide detailed analysis of those phases.

\subsection{Feature Extraction and classification}
For feature extraction, we rely on an existing deep model, ResNet-50 \cite{he2016deep}, pre-trained on ImageNet \cite{deng2009imagenet}. The model is used as feature descriptor without any retraining and fine-tuning. The basic motivation for using the existing pre-trained models as feature descriptor comes from our previous work \cite{ahmad2018comparative,ahmad2019deep1} where we have shown outstanding generalization capabilities on disaster images. Features are extracted from the top fully connected layer resulting in a 1000 dimensional feature vector and the classification phase is based on SVM.   

\subsection{Active Learning}
In this phase, as a first step, we divide the images collected from social media into two sub-sets; namely, (i) initial training set, which is also known as the \textit{seed} and is annotated with human annotators, and (ii) unlabeled pool of images. An SVM classifier is then trained on the initial small labeled training set and the initial accuracy is recorded in the second step. The training set is then populated by querying images from the unlabeled pool of images in step 3, iteratively. To this aim, we employed two methods; namely, (i) Uncertainty Sampling and (ii) Query By Committee. For each method, three different sampling/disagreement strategies are utilized as described in Section \ref{sec:background}. Steps 2 and step 3 are repeated for a given number of iterations as detailed in the experimental setup Section \ref{sec:results}.

\section{Dataset collection}
\label{sec:dataset}

\begin{figure}[!htbp]
\centering
\includegraphics[width=0.9\linewidth]{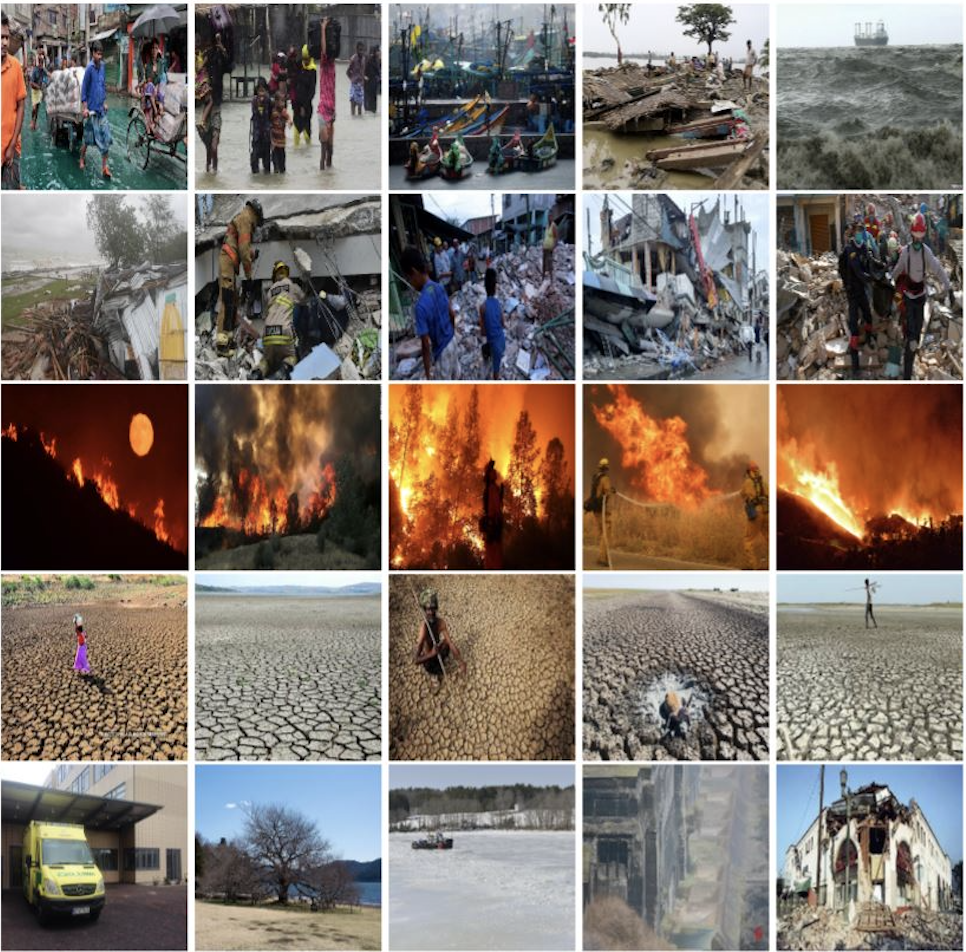}
\caption{Sample images from the dataset.}
		\label{fig:sample_images}
\end{figure}

Our new collected dataset covers images from most common types of natural disasters; including, cyclone, drought, earthquake, floods, landslides, thunderstorm, snowstorm and wildfires. The images are downloaded from social media platforms using the corresponding keywords. The collection of the images is divided into two sub-sets; namely, an initial training set also known as seed and an unlabeled pool of images. For our initial training which is the only part of the training set annotated by human annotators, a subset composed of 160 images collected for each class/type of disaster is randomly selected and annotated by human annotators in a crowd sourcing study. Similarly, the test set, which is composed of 2,516 images, has also been manually examined and annotated in the crowd-sourcing study. The rest of the collected images are treated as an unlabeled pool of images containing a large portion of irrelevant images. Moreover, in the comparison against baselines, for one of the methods as detailed in Section \ref{sec:results}, we also manually annotated the unlabeled pool of images resulting in around 2500 additional annotated images. Figure \ref{fig:sample_images} provides some sample images from the dataset. 

\section{Experimental Setup and Results}
\label{sec:results}

\subsection{Experimental Setup}
The objective of our experiments is manifold. Our objective is to analyze the performance of active learning in support of disaster analysis in images shared on social networks. We also aim to analyze the performance of different active learning techniques when using different sampling/disagreement strategies. Moreover, we want to analyze the difference in the performances of a model/classifier trained on human annotated dataset and training samples collected through the active learning techniques. To achieve those objectives, we performed the following experiments:

\begin{itemize}
	\item First, we analyze the performance of two commonly used techniques of pool-based sampling active learning; namely, uncertainty sampling and query by committee.  
	\item Then, we investigate the impact of using different sampling and disagreement strategies in conjunction with active learning methods on their overall performance.
	\item Finally, we assess and evaluate the performance of active learning techniques against two baseline methods where one of the fully supervised classifiers is trained on labeled data annotated by human annotators while the other is trained on the complete pool of images that includes irrelevant ones.
\end{itemize}

We used the same experimental setup for all our experimental studies. Specifically, our initial training set (seed), annotated manually, is composed of 160 images covering 20 samples from each of the eight different types of natural disasters. Moreover, we used a different number of iterations (max 2000) in our experiments. In each iteration, a single image from the pool of unlabelled images is included in the training set. 

\subsection{Experimental results}

Table \ref{tab:uncertainity_results} provides the evaluation results of the uncertainty sampling method with three different sampling strategies; namely, LC, MS and ES using a variable number of iterations ranging from 1 to 2000 (step size of 250). As expected, the accuracy improves by adding relevant samples from the unlabeled pool of images to the initial training set in each iteration until the accuracy stabilizes for all three methods. Here one important observation is the variation in the performances of the method with the three different sampling techniques as the LC considers only the most probable label, MS considers the top two while ES makes use of all the labels in it decision of choosing a sample from the pool. No significant difference was observed when the number of iterations is around 2000. However, higher variations were observed in the accuracy of the different sampling strategies when the number of iterations is below 1000. At beginning, surprisingly, MS and LC strategies performed well compared to ES, which shows the importance of the make use of most probably labels only in the decision making process. However, relying on the most probably label increases dependence on the accuracy of the initial model/classifier trained on the initial small training set.  

\begin{table}[!htbp]
\begin{center}
\caption{Evaluation of the different sampling strategies for uncertainty sampling based method at different number of queries.}
\label{tab:uncertainity_results}
\scalebox{0.7}{
\begin{tabular}{|c|c|c|c|c|c|c|c|c|c|}
\hline
\multirow{2}{*}{\textbf{Sampling strategy}} & \multicolumn{8}{c|}{\textbf{Accuracy (\%) at different queries}} \\ \cline{2-9} 
& 1 & 250& 500 &750 & 1000& 1250& 1500& 2000 \\ \hline
LC & 0.52& 0.61 & 0.66&0.68 &0.69&0.70 & 0.70& 0.71 \\ \hline
MS & 0.53& 0.65& 0.67&0.67 & 0.69& 0.71& 0.71& 0.71 \\ \hline
ES & 0.52& 0.60 &0.64& 0.66& 0.69&0.68 & 0.69& 0.70 \\ \hline
\end{tabular}}
\end{center}
\end{table}





In Table \ref{tab:query_results}, we provide the experimental results of query by committee based active learning method with different disagreement strategies given a number of iterations. Overall, better accuracy is obtained compared to the uncertainty sampling methods, which is mainly due to employing several hypotheses/models in the sample selection process. As far as the comparison of the disagreement strategies is concerned, slightly better results are observed for the CE and MD strategies compared to the VE.

\begin{table}[]
\begin{center}
\caption{Evaluation results of the different disagreement strategies used for uncertainty sampling at different number of queries.}
\label{tab:query_results}
\scalebox{0.7}{
\begin{tabular}{|c|c|c|c|c|c|c|c|c|}
\hline
\multirow{2}{*}{\textbf{disagreement strategy}} & \multicolumn{8}{c|}{\textbf{Accuracy (\%) at different queries}} \\ \cline{2-9} 
& 1 & 250 & 500 & 750 & 1000& 1250& 1500 & 2000 \\ \hline
VE & 0.49 & 0.65 & 0.69& 0.72& 0.71& 0.69& 0.70& 0.71 \\ \hline
CE & 0.47& 0.66 & 0.67&0.68 & 0.71& 0.71& 0.71& 0.72 \\ \hline
MD & 0.39& 0.64& 0.67& 0.69& 0.71& 0.71& 0.71& 0.72 \\ 
\hline
\end{tabular}}
\end{center}
\end{table}

\begin{figure}[]
\centering
\includegraphics[width=0.95\linewidth]{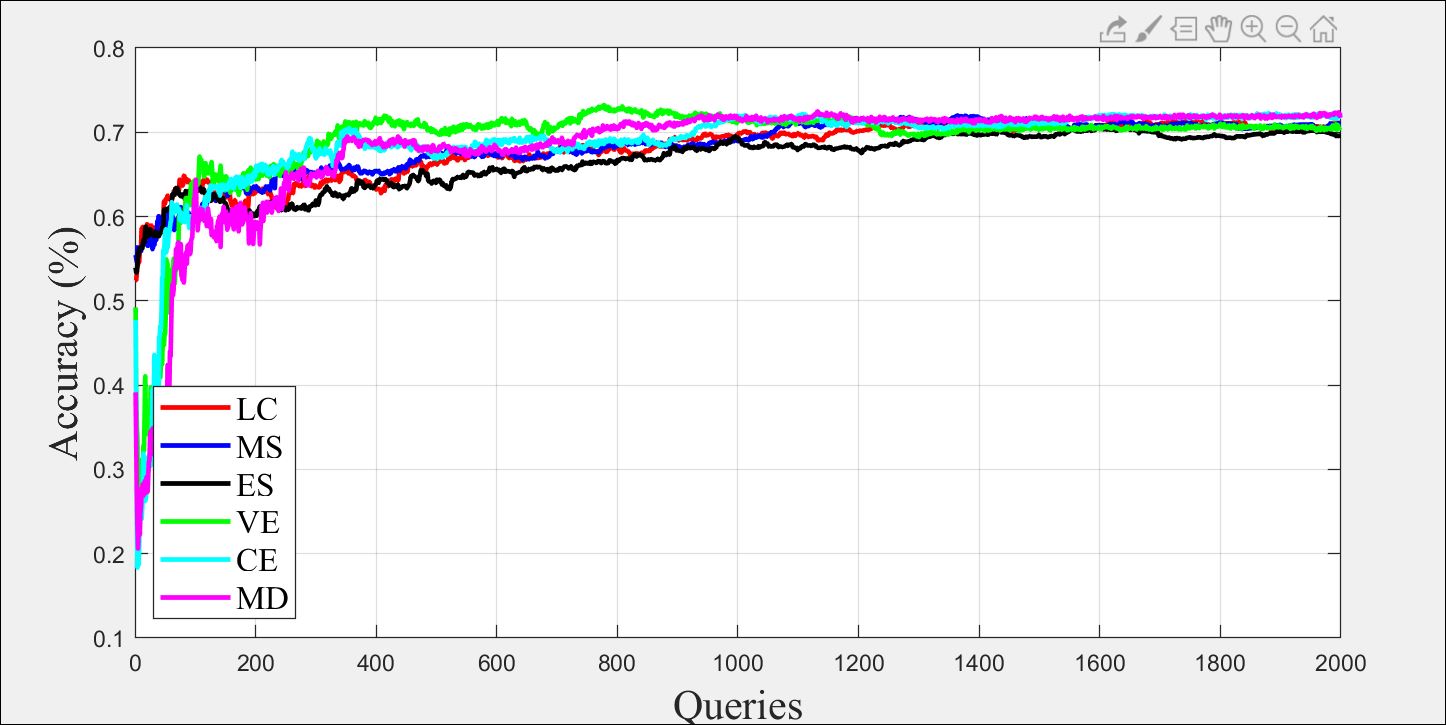}
\caption{Comparison of both methods with different sampling strategies}
	\label{plot2}
\end{figure}

In order to better analyze the variations in the accuracy of these methods with different sampling and disagreement strategies at different iterations, Figure \ref{plot2} provides the performance of the methods with different sampling and disagreement strategies at each iteration. As can be seen, both the methods start at lower accuracy with all sampling and disagreement strategies and improve iteratively. Compared to uncertainty sampling, the cures are more smoother for query by committee method. Moreover, the accuracy improves more rapidly and achieves stability sooner (i.e., after 1000 iterations the accuracy is stabilized).


\vspace{-3mm}

\begin{figure}[]
\centering
\includegraphics[width=0.95\linewidth]{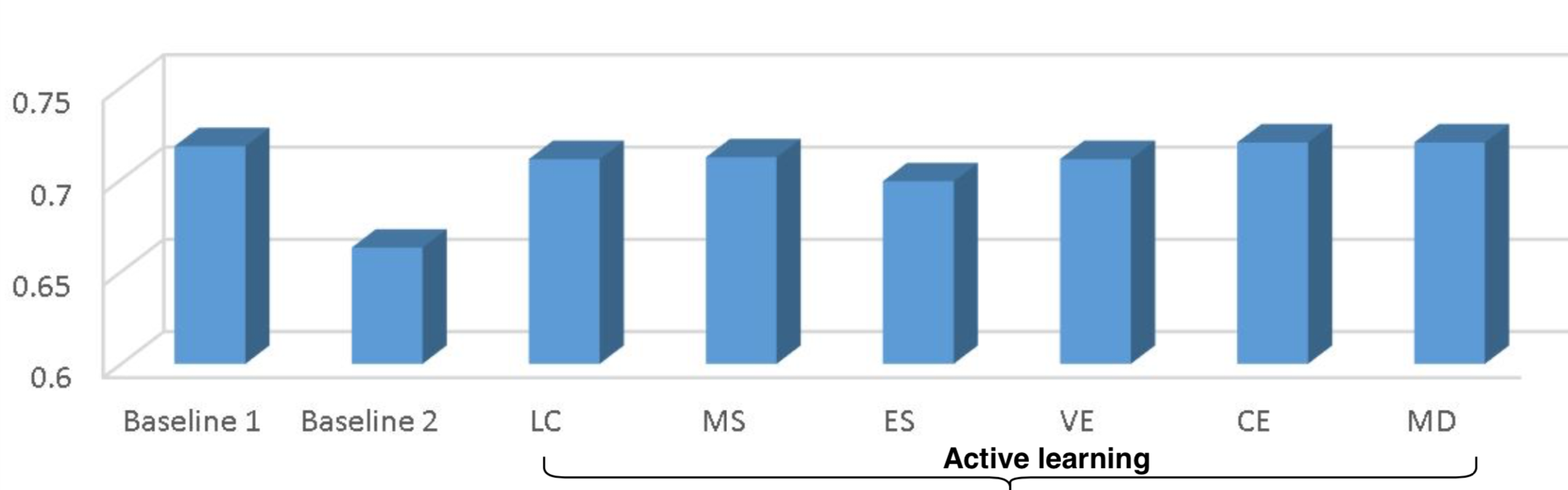}
\caption{Comparisons of the active learning methods against baseline.}
	\label{fig:comparison}
\end{figure}

 The main focus of the paper is to analyze and evaluate the importance/application of active learning techniques in disaster analysis and to show how the active learning component can further improve the performances of disaster analysis frameworks with less annotated data. Thus, in order to show the effectiveness of the active learning methods, instead of sate-of-the-art methods, we compare the results against two extreme cases reported as  baseline 1 and baseline 2 as shown in Figure \ref{fig:comparison}. In the first baseline method, an SVM is trained on human annotated training set, where relevant samples were collected and annotated by human observers from the pool of images. In the experiment, features are extracted with the same deep model (i.e., ResNet) using the same parameters for the SVM classifier. Moreover, a significant amount of training samples (i.e., around 2500) have been used for training the classifier. In the second case, we trained an SVM classifier on the complete pool of images without removing the irrelevant images with the aim to analyze how much the irrelevant images affect the performance of the classifier. 
 As can be seen in most of the cases the active learning methods have comparable results to those obtained from the baseline 1 with fully supervised method, which uses a human annotated training set. Those results illustrate the effectiveness of the active learning techniques where a small annotated dataset is utilized to obtain better results without involving human annotators in the tedious job of annotation large training sets. In the second case, the accuracy has been reduced significantly showing the efficacy of the active learning techniques able to pick right samples for training among the pool of images.  

\subsubsection{Lessons learned}
The lessons learned from the experiments are:

\begin{itemize}
    \item   The  accuracy improves  by  adding  relevant  samples  from  the  unlabeled pool  of  images  to  the  initial  training  set  in  each  iteration until  the  accuracy  stabilizes  at certain point. 
    \item Better accuracy against the baseline methods illustrates the effectiveness of the active  learning  techniques  where  a  small  annotated  dataset is utilized to obtain better results without involving human annotators in the tedious job of annotation large training sets.
\end{itemize}


\section{Conclusion}
\label{sec:conclusion}
In this paper we presented an active learning approach for the disaster analysis in images shared on social media outlets. We mainly used two techniques with several sampling and disagreement strategies for each of the methods. Our experimental results illustrate the effectiveness of using active learning techniques and their ability to produce results comparable to those obtained using human annotated training sets. Our experimental results also illustrate that the classification accuracy improves with the inclusion of images from the unlabelled pool of images in each iteration using active learning. Furthermore, our proposed iterative technique ultimately achieves stability in terms of classification accuracy through the progressive inclusion of images from the unlabelled pool of images. Finally, it has been demonstrated that the query by committee active learning method is more effective for the disaster analysis in images compared to the uncertainty sampling based active learning methods.


\bibliographystyle{spmpsci}      

\bibliography{sigproc}

\end{document}